\pdfoutput=1 
\documentclass[a4paper,twoside]{article}

\usepackage{epsfig}
\usepackage{subcaption}
\usepackage{calc}
\usepackage{amssymb}
\usepackage{amstext}
\usepackage{amsmath}
\usepackage{amsthm}
\usepackage{multicol}
\usepackage{pslatex}
\usepackage{apalike}

\usepackage{tabularx}
\usepackage{float}
\usepackage{url}
\usepackage{xspace}
\usepackage{xstring}

\usepackage{SCITEPRESS}

\newcolumntype{Y}{>{\centering\arraybackslash}X}

\newcommand{\citationneeded}[1][]{\textsuperscript{[citation needed]}}
\newcommand{\subu}[1]{\StrSubstitute{#1}{_}{ }}

\def\imgsize{0.3\textwidth}

\newcommand{\specificresults}[6]{
\begin{figure}[H]
\captionsetup[subfigure]{justification=centering}
\begin{subfigure}{\imgsize}
  \centering
  \includegraphics[width=.8\textwidth]{results/specific/#1/#3.jpg}
  \caption{\subu{#3}}
\end{subfigure}
\hfill
\begin{subfigure}{\imgsize}
  \centering
  \includegraphics[width=.8\textwidth]{results/specific/#1/#4.jpg}
  \caption{\subu{#4}}
\end{subfigure}
\hfill
\begin{subfigure}{\imgsize}
  \centering
  \includegraphics[width=.8\textwidth]{results/specific/#1/#5.jpg}
  \caption{\subu{#5}}
\end{subfigure}
\vspace{0.3cm}
\caption{#2}
\label{#6}
\end{figure}
}

\newcommand{\mixresults}[2]{

\begin{figure}[H]
\captionsetup[subfigure]{justification=centering}
\begin{subfigure}[t]{\imgsize}
  \centering
  \includegraphics[width=.8\textwidth]{results/mix/#1/StyleGAN2_ffhq_d.jpg}
  \caption{StyleGAN2-face}
\end{subfigure}
\hfill
\begin{subfigure}[t]{\imgsize}
  \centering
  \includegraphics[width=.8\textwidth]{results/mix/#1/StyleGAN2_car_d.jpg}
  \caption{StyleGAN2-car}
\end{subfigure}
\hfill
\begin{subfigure}[t]{\imgsize}
  \centering
  \includegraphics[width=.8\textwidth]{results/mix/#1/StyleGAN2_church_d.jpg}
  \caption{StyleGAN2-church}
\end{subfigure}

\vspace{0.5cm}

\begin{subfigure}[t]{\imgsize}
  \centering
  \includegraphics[width=.8\textwidth]{results/mix/#1/StyleGAN2_ffhq_nod.jpg}
  \caption{StyleGAN2-face*}
\end{subfigure}
\hfill
\begin{subfigure}[t]{\imgsize}
  \centering
  \includegraphics[width=.8\textwidth]{results/mix/#1/StyleGAN2_car_nod.jpg}
  \caption{StyleGAN2-car*}
\end{subfigure}
\hfill
\begin{subfigure}[t]{\imgsize}
  \centering
  \includegraphics[width=.8\textwidth]{results/mix/#1/StyleGAN2_church_nod.jpg}
  \caption{StyleGAN2-church*}
\end{subfigure}
\vspace{0.3cm}
\caption{Output images generated via StyleGAN2 - with and without (*) Discriminator - for the input caption ``\subu{#1}''.}
\label{#2}
\end{figure}
}

\newcommand{\gptresults}{

\begin{figure}[H]
\captionsetup[subfigure]{justification=centering}
\begin{subfigure}[t]{\imgsize}
  \centering
  \includegraphics[width=.8\textwidth]{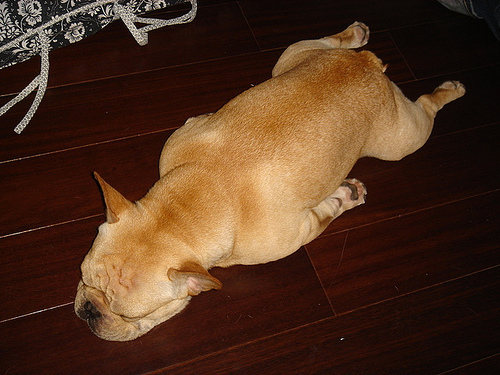}
  \caption{the picture of a dog that has a high fatality rate}
\end{subfigure}
\hfill
\begin{subfigure}[t]{\imgsize}
  \centering
  \includegraphics[width=.8\textwidth]{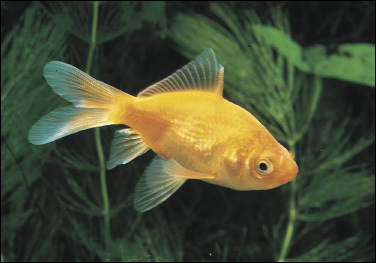}
  \caption{the picture of the fish.}
\end{subfigure}
\hfill
\begin{subfigure}[t]{\imgsize}
  \centering
  \includegraphics[width=.8\textwidth]{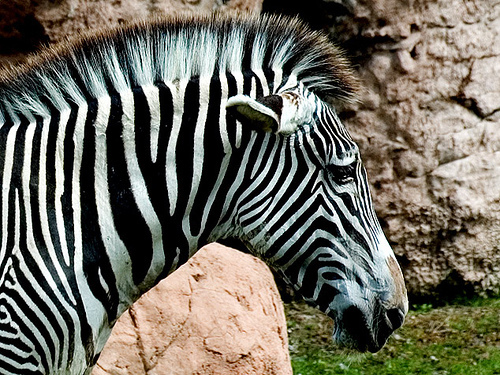}
  \caption{the picture of a zebra in a zebra coat.}
\end{subfigure}

\vspace{0.5cm}

\begin{subfigure}[t]{\imgsize}
  \centering
  \includegraphics[width=.8\textwidth]{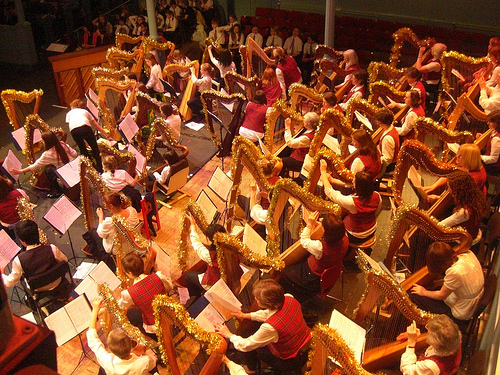}
  \caption{the picture of the orchestra}
\end{subfigure}
\hfill
\begin{subfigure}[t]{\imgsize}
  \centering
  \includegraphics[width=.8\textwidth]{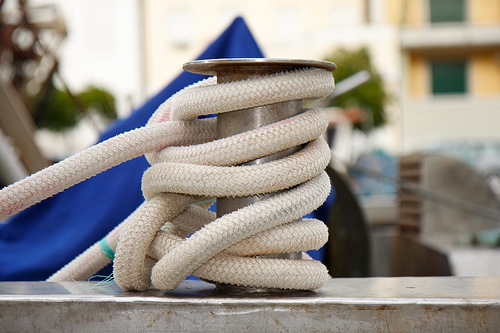}
  \caption{the picture of the art of knots}
\end{subfigure}
\hfill
\begin{subfigure}[t]{\imgsize}
  \centering
  \includegraphics[width=.8\textwidth]{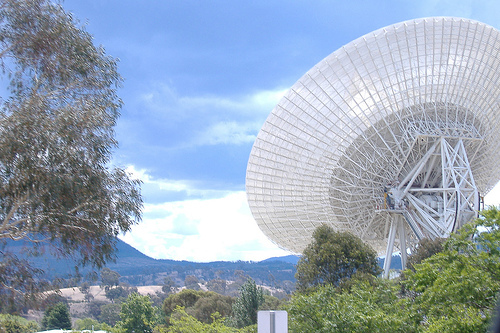}
  \caption{the picture of the world's largest radio telescope}
\end{subfigure}

\vspace{0.5cm}

\begin{subfigure}[t]{\imgsize}
  \centering
  \includegraphics[width=.8\textwidth]{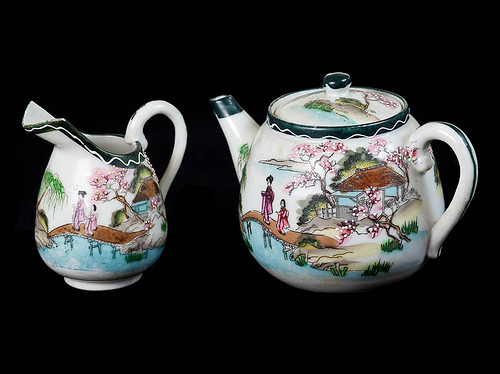}
  \caption{the picture of the pottery of the city of Suzhou}
\end{subfigure}
\hfill
\begin{subfigure}[t]{\imgsize}
  \centering
  \includegraphics[width=.8\textwidth]{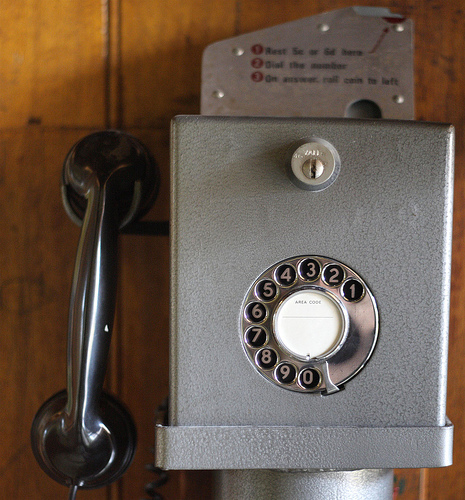}
  \caption{the picture of the world's first ``safe'' phone}
\end{subfigure}
\hfill
\begin{subfigure}[t]{\imgsize}
  \centering
  \includegraphics[width=.8\textwidth]{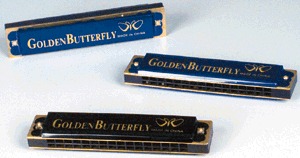}
  \caption{the picture of the butterfly package}
\end{subfigure}

\vspace{0.3cm}
\caption{Input images and related output captions generated via GPT2}
\label{fig:results_gpt2}
\end{figure}
}

\newcommand{\glass}{CLIP-GLaSS\xspace}

\usepackage{pdflscape}
\usepackage{pdfpages}
\makeatletter
\@ifundefined{@setmarks}{\let\@setmarks\relax}{}
\makeatother
\newcommand{\citationpage}[1]{
    \begin{landscape}
    \includepdf[pages=-, angle=90]{#1}
    \end{landscape}
}

\begin{document}

\citationpage{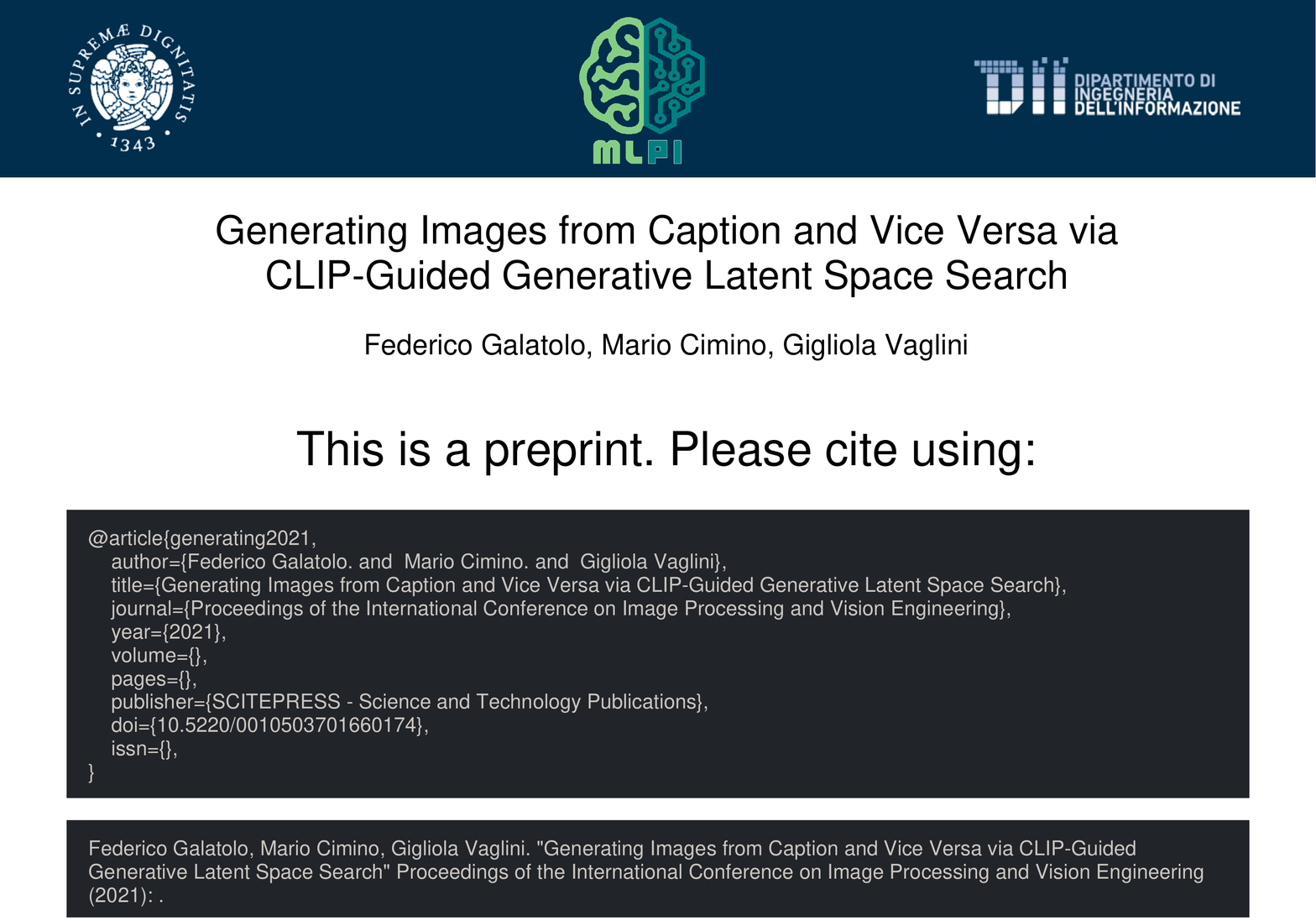}

\title{Generating images from caption and vice versa \\via CLIP-Guided Generative Latent Space Search}

\author{\authorname{Federico A. Galatolo\sup{\ast 1}\orcidAuthor{0000-0001-7193-3754}, Mario G.C.A. Cimino\sup{1}\orcidAuthor{0000-0002-1031-1959} and Gigliola Vaglini\sup{1}\orcidAuthor{0000-0003-1949-6504}}
\affiliation{\sup{1}Department of Information Engineering, University of Pisa, 56122 Pisa, Italy}
\email{\sup{\ast}federico.galatolo@ing.unipi.it}
}


\abstract{In this research work we present \glass, a novel zero-shot framework to generate an image (or a caption) corresponding to a given caption (or image). \glass is based on the CLIP neural network, which, given an image and a descriptive caption, provides similar embeddings. Differently, \glass takes a caption (or an image) as an input, and generates the image (or the caption) whose CLIP embedding is the most similar to the input one. This optimal image (or caption) is produced via a generative network, after an exploration by a genetic algorithm. Promising results are shown, based on the experimentation of the image Generators BigGAN and StyleGAN2, and of the text Generator GPT2.
}

\onecolumn \maketitle \normalsize \setcounter{footnote}{0} \vfill

\section{\uppercase{Introduction and background}}

In the last years, Generative Neural Networks showed promising results in a wide range of fields. The state of the art in the field of image generation is represented by Generative Adversarial Networks (GANs) \cite{gan}. GANs are capable of generating realistic synthetic images leveraging two competing neural networks: a Generator and a Discriminator. The objective of the Generator is to generate images capable of fooling the Discriminator. The objective of the Discriminator is to distinguish between images generated by the Generator and the original ones.\\
In the field of Natural Language Processing (NLP), unprecedented results have been achieved by transformers architectures \cite{attentionsurvey}. One of the most known and studied transformer is the Generative Pre-trained Transformer 2 (GPT2) \cite{gpt2}. GPT-2 has 1.5 billion parameters and was trained on a language modelling task on the texts of 8 millions of web pages.\\
Generating images from a descriptive caption has always been a challenging problem. In the last years, some architectures like StackGAN++ \cite{stackgan++} and AlignDRAW \cite{attention_captions} showed promising results in this field, although being limited to the visual and textual domains of the training dataset.
Very recently (January 2021), a novel deep network which learns visual concepts from natural language supervision has been released by OpenAI. CLIP (Contrastive Language–Image Pre-training) \cite{CLIP} consists of two encoders: one for images and another for texts. CLIP encoders are capable of producing similar embeddings for images and texts representing similar concepts. CLIP can be applied to visual classification tasks without training: to distinguish the object X from Y in an images dataset, it is sufficient for each image to check whether the text description “a photo of X” or “a photo of Y” is more likely to be paired with it.\\
In this paper, we propose a framework based on CLIP to generate (i.e., to build without a supporting database) the best image corresponding to a target caption. The proposed framework can also be used to generate a caption corresponding to a given image. More specifically, the framework takes a caption (or an image) as an input, and generates the image (or the caption) whose CLIP embedding is most similar to the input one. This optimal image (or text) is produced via a generative network after an exploration by a genetic algorithm. Early experimentation of the proposed CLIP-guided Generative Latent Space Search (\glass) has been carried out, on the basis of the image Generators BigGAN and StyleGAN2, and of the text Generator GPT2, for the text-to-image and image-to-text tasks, respectively.

The paper is structured as follows. Section 2 focuses on the Design of the \glass framework. Experimental studies are covered by Section 3. Conclusions are drawn in Section 4. The source code of \glass has been publicly released. 

\section{DESIGN OF THE \glass FRAMEWORK}
The main components of the \glass framework are: (i) the CLIP network for producing image and text embedding, (ii) an Image or Text Generator for generating a text/image output with a similar embedding, (iii) a Genetic Algorithm to explore the latent space of the Generator for finding the most similar embedding between image and text. In the case of the text-to-image task, the Image Generator is based on domain-specific or mixed-domains Generative Adversarial Networks, such as DeepMind's BigGAN and Nvidia's StyleGAN2, respectively. In the case of the image-to-text task, the Generative Pre-trained Transformer 2 (GPT2) has been used. Finally, as a Genetic Algorithm, the NSGA-II \cite{nsga2} has been employed, in order to solve a multi objective optimization problem. A classical Genetic Algorithm can be employed when solving one single objective optimization. The advantage of the Genetic Algorithm is that it is independent from the type of generative architecture, and it is characterized by a high degree of exploration. Since the genetic algorithm is considered well-known, the next subsections detail the first two components.

\subsection{The CLIP network}
CLIP is composed by two Neural Networks: an Image Encoder (IE) and a Text Encoder (TE). The two encoders produce similar embeddings if the image and the text contains similar visual and textual concepts. CLIP was trained using images and related snipped of texts scraped from the internet, to perform a contrastive learning task. Contrastive learning consists in training a model to predict the correct similarity between data samples. CLIP was inspired by some prior work: in 2013 Richer Socher \textit{et al.} trained a model to map images in feature vectors close to semantic word vectors corresponding to their class. The model shown some zero-shot capabilities \cite{socher2013zero}. Later, in 2017, Ang Li \textit{et al.} trained a model using images scraped form the internet and related user comments to predict the corresponding comments word n-gram given an image. The resulting model was able to achieve a zero-shot 11.5\% accuracy on ImageNet \cite{li2017learning}.
Because of the wide range of visual concepts learned from natural language via contrastive learning, CLIP shows great zero-shot capabilities. CLIP's zero-shot performances were tested on over 30 different tasks spanning from OCR to geo-localization. The model showed competitive results against fully supervised baselines. CLIP was able to match the accuracy of the original ResNet-50 classifier on ImageNet in zero-shot without seeing any of the 1.28 million training images.

\subsection{Design of the Text-to-Image task}

\begin{figure}[h]
\includegraphics[width=0.47\textwidth]{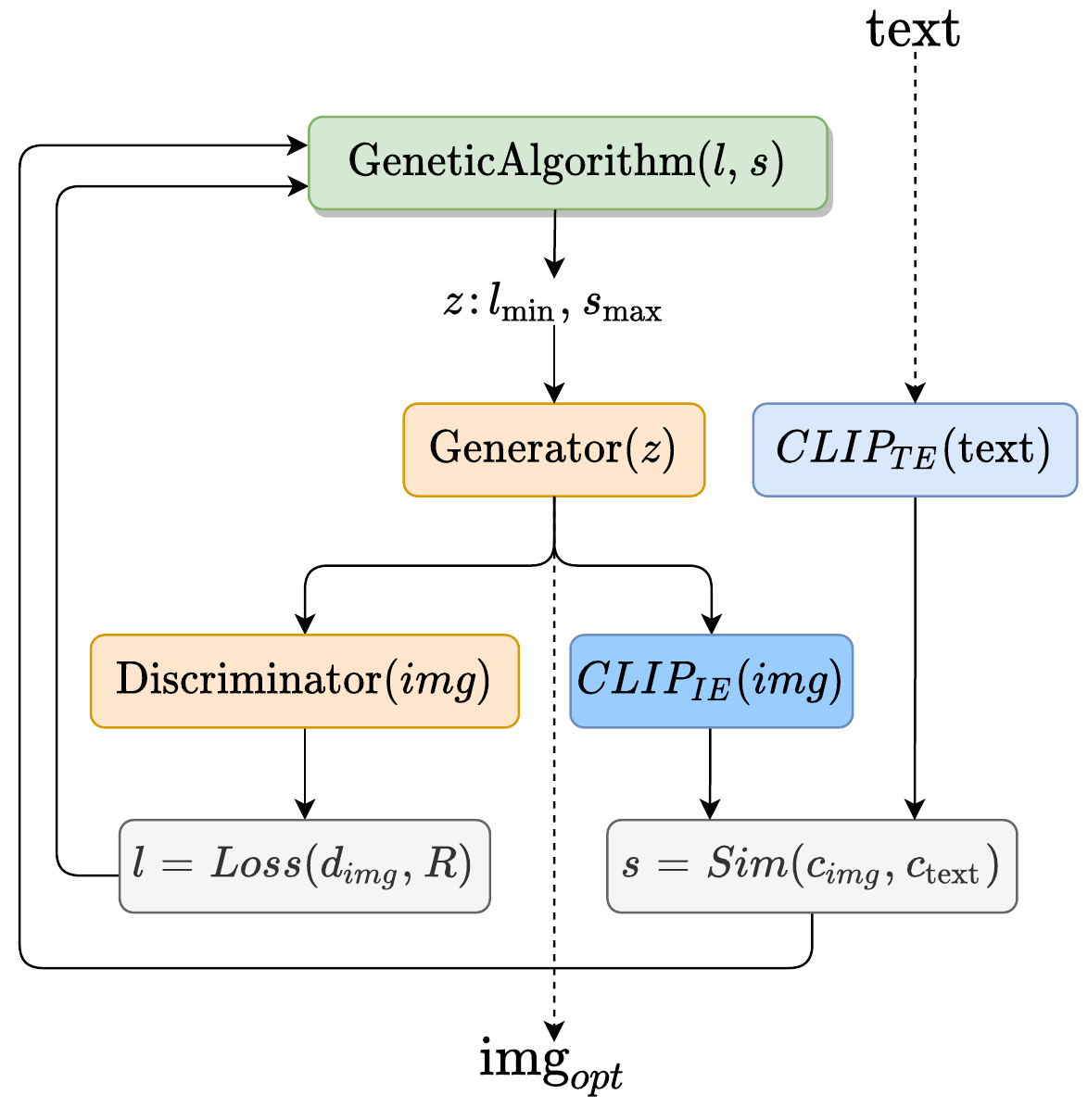}
\caption{Architectural design of the \glass framework for the text-to-image task} \label{fig:txt2img}
\end{figure}

Figure \ref{fig:txt2img} shows the architectural design of the \glass framework for the text-to-image task. Here, the CLIP text/image embeddings are represented in light/dark blue, respectively. The similarity $s$ of their output is sent to the Genetic Algorithm (the green box on the top) for being maximized. The Genetic Algorithms controls the input $z$ of the image Generator, whose components are represented in orange. The image Generator is based on a Generative Adversarial Network (GAN). To clearly understand the overall behavior of the framework, the GAN is first introduced in the following.

The Generative Adversarial Neural Network is a framework proposed in 2014 \cite{gan}, and consists in two competing neural networks: a $Generator$ and a $Discriminator$. The Generator produces candidates while the Discriminator evaluates them. The Generator learns to map from an seeding space called \textit{latent space} ($z$), to the data space of interest, while the Discriminator distinguishes candidates produced by the Generator ($d_{img}$) from the real data ($R$). In Figure~\ref{fig:txt2img}, the output of the Discriminator is evaluated in terms of loss ($l$) to be minimized. During the training, the Generator objective is to increase the error rate of the Discriminator, by producing novel candidates that the Discriminator classifies as real. For this purpose, it is seeded with randomized input that is sampled from a predefined latent space (noise distribution $p_z$). Independent backpropagation procedures are applied to both networks, to allow the Generator to produce better images, while the Discriminator becomes more skilled at distinguishing synthetic images. For this purpose, Generator and Discriminator are simultaneously trained to perform their respective task in a zero-sum game. The Generator and the Discriminator are typically transposed convolutional and convolutional neural networks, respectively. In the proposed architecture, pre-trained GANs have been used. Specifically, the Discriminator takes the image and provides a single scalar representing the probability that its input comes from the original data rather than from the Generator. More formally, let us assume that the Discriminator $D$ is trained to output $1$ if its input image is original, and $0$ if it is synthetically generated. Then, using the Cross Entropy loss the conjunct training objective function can be expressed as:
\begin{align*}
    &\min\limits_{G} \max\limits_{D} \mathbb{E}_{x \sim data}[log(D(x)]\;+
\end{align*}
\begin{align}
    &\mathbb{E}_{z \sim p_z}[log(1 - D(G(z)))]
\end{align}
where $\mathbb{E}_{x \sim p}$ means the expectation over the probability distribution $p$.

At the end of the training process, the Generator is able to generate data indistinguishable (by the Discriminator) from the data originally provided. It has been shown that the resulting Generator is also able to give semantically significant interpolations in the domain of the training data \cite{gansurvey}.
Figure \ref{fig:txt2img} focuses on the search carried out by the Genetic Algorithm, over pre-trained networks. Overall, the Generator is seeded by $z$ (according to a noise distribution $p_z$) and provides a related image to both the Discriminator and the CLIP image encoding ($CLIP_{IE}$). The latter is compared with the CLIP text encoding ($CLIP_{TE}$) of the target text, via a similarity function $Sim$. As a similarity function, the cosine similarity has been used according to \cite{CLIP}.  One objective for the Genetic Algorithm is to provide the best $z$ to maximize this similarity. More formally, given the target text $T$, the optimization problem is:
\begin{align}
    \max\limits_{z}\;\;sim(CLIP_{IE}(G(z)),\;CLIP_{TE}(T)). 
\end{align}
The second objective of the Genetic Algorithm is the classification loss $l$ of the Discriminator, which is calculated from the encoding $d_{img}$ provided by the image of the Generator, and the value associated to the output of a real image ($R$). More formally, the optimization problem is: 
\begin{align}
    \min\limits_{z}\;\;Loss(D(G(z)), R).
\end{align}

After solving the optimization problem, the resulting optimal image provided by the Generator is $img_{opt}$.\\
It is worth to notice that if using a generative architecture without Discriminator, the second objective is missing. As a consequence, the optimization problem is single-objective, since it considers only the CLIP embeddings similarity.

\subsection{Design of the Image-to-Text task}

\begin{figure}[h]
\includegraphics[width=0.5\textwidth]{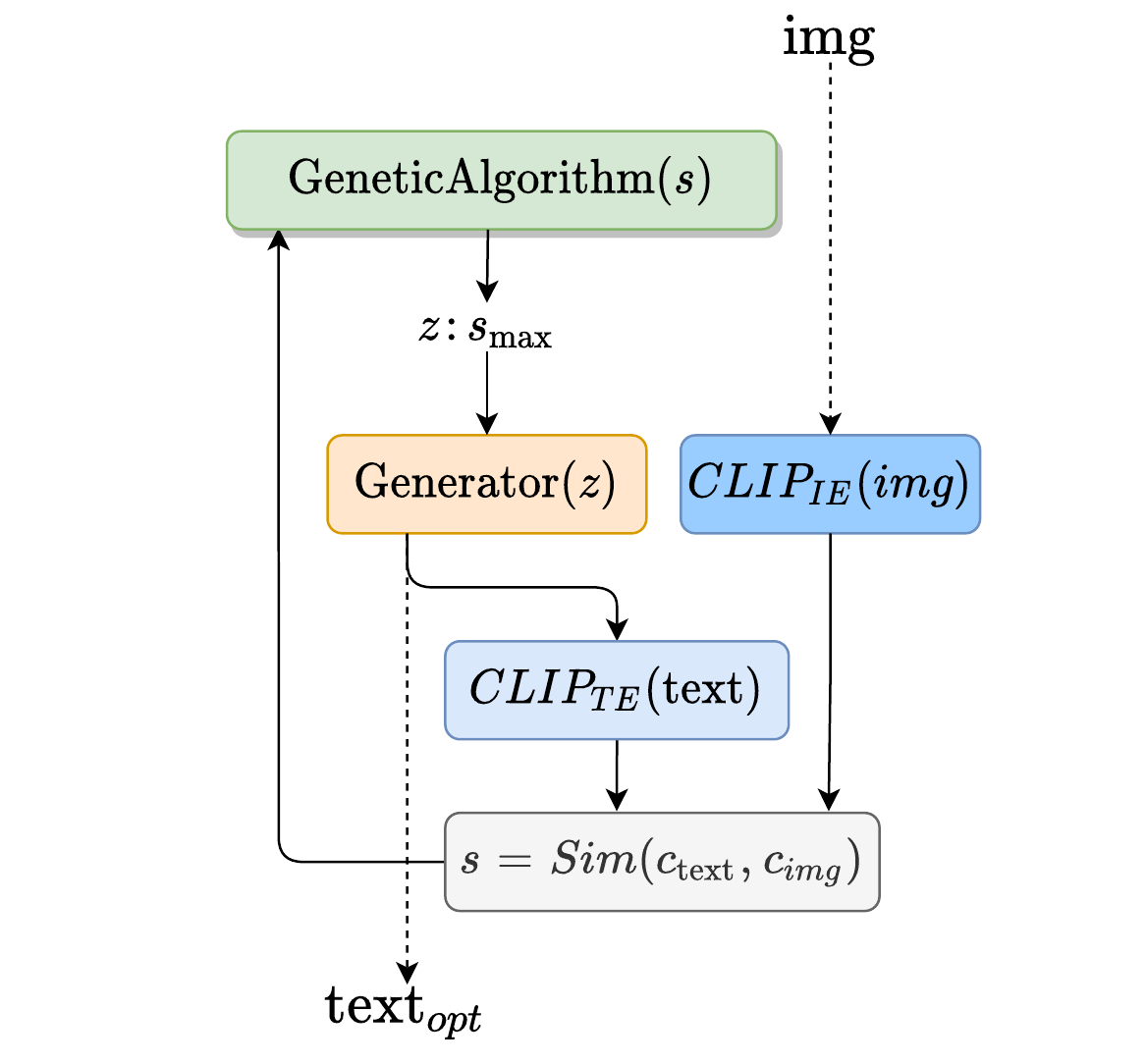}
\caption{Architectural design of the \glass framework for the image-to-text task} \label{fig:img2txt}
\end{figure}

Figure \ref{fig:img2txt} shows the architectural design of the \glass framework for the image-to-text task. Similarly to the text-to-image task, the CLIP image/text embeddings are represented in dark/light blue, respectively. The similarity $s$ of their output is sent to the Genetic Algorithm for being maximized. The Genetic Algorithms controls the seed $z$ of the text Generator, represented in orange. The state of the art for text generators is based on transformers, such as XLNet \cite{xlnet}, Conditional Transformer Language Model (CTRL)\cite{ctrl} and Generative Pre-trained Transformer-2 (GPT2)\cite{gpt2}. As an example, in this paper the GPT2 has been used.

More specifically, GPT2 is a transformer model developed by OpenAI as improved version of the Generative Pre-trained Transformer (GPT). The transformer architecture was introduced in 2017, and it is used mainly for solving NLP tasks. Although transformers are built to work with temporal data (like Recurrent Neural Networks or Long Short Term Memory Neural Networks), they do not require that temporal data are sequentially processed. Hence, transformers allow a fast large scale parallel training.\\
Transformers are composed by a stack of encoders and decoders, and heavily rely on a mechanism called \textit{attention}, which is able to highlight important information while discarding the non-important one. The training data for GPT2 was scraped from the internet using a custom web scraper that emphasizes document quality using some meta-heuristics. The resulting dataset contains text from over 45 million links and consists in almost 40 GB of text. GPT2 was trained using a Model-Agnostic Meta-Learning (MAML) method \cite{maml}, and tested with zero-shot approach on a variety of tasks: text generation, translation, summarization, question answering, etc. It was able to match and even surpass the state-of-the-art of fully supervised models in zero-shot.\\
It is important to notice that GPT2 does not use a human-like alphabet, but an alphabet of tokens generated using the Byte Pair Encoding (BPE) \cite{bpe}. GPT2 can be used as generative architecture by setting context tokens, and using them to predict the next token, until the stop-sentence token predicted or a target text length is reached. We will refer to input context tokens as its \textit{latent space}. 

In Figure \ref{fig:img2txt} the Generator is seeded by $z$, to provide an output text accordingly. The output text is used to fed the CLIP text encoder ($CLIP_{TE}$). Finally, the similarity between the embedding of the ($CLIP_{TE}$) and the embedding of the CLIP image embedding ($CLIP_{IE}$) is computed. The optimization problem of the Genetic Algorithm is to maximize this similarity.\\
After solving the optimization problem, the resulting optimal image generated from GPT2 is $text_{opt}$.\\

\section{\uppercase{Experimental studies}}
The \glass framework has been implemented and publicly released as an open source GitHub repository \cite{clip-glass-repo}, along with an in-browser demonstration. Experiments have been carried out on an Intel Core i9-9900K CPU, a GeForce RTX 2080 GPU. After the input image/caption, 500 generations are executed by the Genetic Algorithm to find the optimal caption/image. The order of magnitude of the processing time of an input is 5-10 minutes depending on the generative architecture used.

In this section, some pilot experiments of \glass are considered, to show its generation capabilities. Each  output is assessed in terms of $quality$, i.e. absence of artifacts, and $relevance$, i.e., the absence of unexpected elements, evaluated with the naked eye as low, medium, or high.
\subsection{Examples of the Text-to-Image task}
Different state-of-the-art pre-trained networks have been used as Generator/Discriminator. In particular, two GANs have been used: DeepMind's BigGAN \cite{biggan} and Nvidia's StyleGAN2 \cite{stylegan2}.
The original BigGAN, trained on the ImageNet  dataset, has been considered \cite{imagenet}. Three versions of StyleGAN2 have been used: (i) StyleGAN2-face, which is trained on Flickr-Faces-HQ \cite{ffhq}; (ii) StyleGAN2-church, trained on a subset of LSUN \cite{lsun} made of church buildings; (iii) StyleGAN2-car, trained on a subset of LSUN made of cars. \\
OpenAI publicly released only BigGAN Generator. For this case, a single-objective genetic algorithm is used when optimizing its latent space. Differently, Nvidia released both StyleGAN Generator and Discriminator.\\
The BigGAN latent space $z$ is composed by 1000 booleans representing the 1000 ImageNet classes, and by 128 real numbers meant to be sampled from a truncated normal distribution in the range $[-2,\;2]$. When optimizing its latent space, mixed genetic operators are employed to correctly perform the initialization, mutation and crossover operations.\\
The StyleGAN2 latent space $z$ is made by 512 real numbers meant to be sampled from a normal distribution.\\
Figure~\ref{fig:results_specific_StyleGAN2_ffhq_d} shows three representative examples of input caption and related output image generated via StyleGAN2-face. Since this GAN is specialized on faces, the overall result is very good: the quality and the relevance of all images are high, except for Image (b), whose relevance is medium due to the blonde hairs on the bottom.\\
Figure~\ref{fig:results_specific_StyleGAN2_car_d} shows three representative examples of input caption and related output image generated via StyleGAN2-car. Although this GAN is specialized on cars, the quality of all images is medium, due to the presence of artifacts. The relevance is high for (a) and (b), but medium for (c) because the "intersection" is not visible.\\
Figure~\ref{fig:results_specific_StyleGAN2_church_d} shows three representative examples of input caption and related output image generated via StyleGAN2-church. This GAN is specialized on images of church buildings. Indeed, the image relevance is high, and the quality is about high due to the presence of minor artifacts.\\
The examples clearly show that \glass is able to combine the right image elements for matching the target caption, when using input texts that belong to the same domain of the generative network.

Differently than in the previous cases, Figure~\ref{fig:results_mix_ffhq} shows the output images generated via StyleGAN2 on three domains (face, car, and church). To asses the role of the Discriminator, the optimization is also performed without it. In Figure~\ref{fig:results_mix_ffhq}, the images produced without Discriminator have a final asterisk in the caption. Specifically, by using the input text "the face of a blonde girl with glasses", the StyleGAN2-face achieves a high quality and relevance for (a) and (d), as expected. On the other side, the low performance of StyleGAn2-car and StyleGAn2-church are apparent. However, it is worth noting that in Figure~\ref{fig:results_mix_ffhq} (f) the picture resembles a face with glasses, generated via two windows on a building.

Figure~\ref{fig:results_mix_car} and Figure~\ref{fig:results_mix_church} show the output images generated via StyleGAN2, with and without Discriminator, for the input caption "a blue car in the snow" and "a gothic church in the city", respectively. Not surprisingly, the GAN that is specialized in the category of interest (face, car, church) provide the best significance, but medium-low quality due to the presence of artifacts.

Overall, the resulting images resemble the target caption, but the medium-low quality of the images suggests that, to correctly perform this task, a bigger Generator (i.e. with more parameters) trained on a wider variety of images is needed. In other words, the Generator cannot create images that do not belong to its training domain. In general, it is apparent that the Discriminator guarantees a better quality of the output images. Interestingly, when the Discriminator is not used, even if the target text is outside of the Generator domain, it is able to generate images that somehow resemble the target text. For example, Figure \ref{fig:results_mix_car}(d) is generated via StyleGAN2-face with the input caption ``a blue car in the snow'': it represents a face of man in the snow with a blue sweater. Another example is Figure 
 \ref{fig:results_mix_car}(f), generated by StyleGAN2-church without Discriminator: it represents a church with a blue blob whose shape remembers a car.
Figure \ref{fig:results_mix_church} shows examples generated by the three domain StyleGAN2 networks via the caption ``a gothic church in the city''. Specifically, StyleGAN2-car without Discriminator generates an image with a background building that remember the gothic style.
Finally, the \glass framework has been experimented using BigGAN, a large scale GAN trained with ImageNet, i.e., with multiple domains images. Figure~\ref{fig:results_specific_DeepMindBigGAN} shows three captions and the related images generated from BigGAN. Although the overall quality is medium for the presence of artifacts, the relevance is high.

\subsection{Examples of the Image-to-Text task}
This section focuses on some representative examples of the caption generation capabilities of \glass. 
Specifically, 20 context tokens have been used as GPT2 latent space $z$, to which three fixed tokens have been concatenated, representing the static context "the picture of". The latent space of the context tokens is a integer space of numbers ranging from 0 to 50257, which is the BPE vocabulary size. Indeed, GPT2 adopts a subword-level vocabulary: it does not predict the next word but the next subword token. When optimizing GPT2 latent space, integer genetic operators have been used to perform initialization, mutation and crossover operations.\\
Figure~\ref{fig:results_gpt2} shows nine input images randomly extracted from ImageNet, with the related captions generated via GPT2. The results clearly show the \glass potential of caption generation. The most captions have both high quality and high relevance. Some captions, e.g., (a), (c), and (h) have high relevance and medium quality because of the textual artifacts. For example, caption (a) ``the picture of a dog that has a high fatality rate'', can be related to the fact that the dog is lying on the ground; caption (h) ``the picture of the world's first `safe' phone'', can be related to the fact that the phone in the picture resembles a safe.

\section{\uppercase{Conclusions}}
In this research work we have introduced \glass, a zero-shot framework which takes an input caption and generates a corresponding image, and vice versa. \glass is based on the CLIP neural network, for generating close embeddings of semantically close texts or images, a Generator network, for controlling the respective generation of images or texts, and a Genetic Algorithm, to explore the Generator space to find the best image or text. The design choices are first detailed, and then a set of pilot experiments are discussed, using the generative networks BigGAN, StyleGAN2 and GPT2. Results show the high potential of the proposed framework, in terms of quality and relevance of the output image or text, encouraging further comparative research. The source code has been publicly released, along with an in-browser demonstration. 

\section*{\uppercase{Acknowledgement}}
Work partially supported by the Italian Ministry of Education and Research (MIUR) in the framework of the CrossLab project (Departments of Excellence).

\bibliographystyle{apalike}
{\small
\bibliography{bibliography}}

\newpage
\onecolumn

\specificresults{StyleGAN2_ffhq_d}{Input caption and related output image generated via StyleGAN2-face.}{the_face_of_a_man_with_brown_eyes_and_stubble_beard}{the_face_of_a_blonde_woman_with_blue_eyes_and_glasses}{the_face_of_a_bald_man_with_beard_and_brown_eyes}{fig:results_specific_StyleGAN2_ffhq_d}

\specificresults{StyleGAN2_car_d}{Input caption and related output image generated via StyleGAN2-car.} {a_black_SUV_in_the_snow}{a_red_truck_in_a_street}{a_yellow_beetle_in_an_intersection}{fig:results_specific_StyleGAN2_car_d}

\specificresults{StyleGAN2_church_d}{Input caption and related output image generated via StyleGAN2-church.} {a_gothic_church_a_grass_field}{an_orthodox_church_in_moscow}{a_basilica_in_rome}{fig:results_specific_StyleGAN2_church_d}

\mixresults{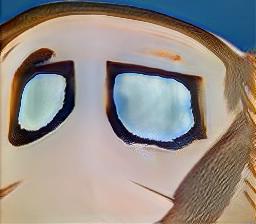}{fig:results_mix_ffhq}
\mixresults{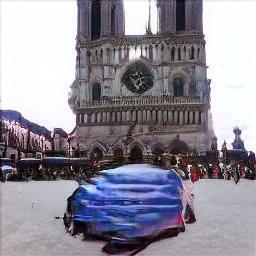}{fig:results_mix_car}
\mixresults{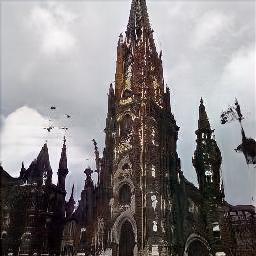}{fig:results_mix_church}

\specificresults{DeepMindBigGAN}{Input caption and related output image generated via BigGAN}{a_clock_in_a_wall}{a_dog_in_the_woods}{a_mountain_lake}{fig:results_specific_DeepMindBigGAN}

\gptresults
\twocolumn

\end{document}